\documentclass{svproc}
\usepackage{graphicx}
\usepackage{balance}
\usepackage{amssymb}
\usepackage{epstopdf}
\usepackage{listings}
\usepackage{amsmath}
\usepackage{bigstrut}
\usepackage{tabularx}
\usepackage{booktabs}
\usepackage[font=small,skip=0pt]{caption}

\usepackage{tikz}
\usepackage{pgfplots}
\usepackage{filecontents}
\pgfplotsset{width=8cm,compat=1.9}
\usetikzlibrary{pgfplots.dateplot}
\usepackage{pgfplotstable}

\usepackage[ruled,vlined,linesnumbered,lined,commentsnumbered]{algorithm2e}

\usepackage{url}

\begin{document}
	\mainmatter              
	\title{Misogynistic Tweet Detection: Modelling CNN with Small Datasets}
	\titlerunning{Misogynistic Tweet Detection}  
	%
	\author{Md Abul Bashar \inst{1} \and Richi Nayak \inst{1}
		\and Nicolas Suzor \inst{2} \and Bridget Weir \inst{2}}
	\authorrunning{Bashar et al.} 
	%
	%
	\institute{School of Electrical Engineering and Computer Science\\
		\and
		School of Law\\
		Queensland University of Technology, 
		Brisbane, Australia\\
		\email{\{m1.bashar, r.nayak, n.suzor, bridget.weir\}@qut.edu.au}
	}
	
	\maketitle              
	\vspace{-0.5cm}
	\begin{abstract}
		Online abuse directed towards women on the social media platform Twitter has attracted considerable attention in recent years. An automated method to effectively identify misogynistic abuse could improve our understanding of the patterns, driving factors, and effectiveness of responses associated with abusive tweets over a sustained time period. However, training a neural network (NN) model with a small set of labelled data to detect misogynistic tweets is difficult. This is partly due to the complex nature of tweets which contain misogynistic content, and the vast number of parameters needed to be learned in a NN model. We have conducted a series of experiments to investigate how to train a NN model to detect misogynistic tweets effectively. In particular, we have customised and regularised a Convolutional Neural Network (CNN) architecture and shown that the word vectors pre-trained on a task-specific domain can be used to train a CNN model effectively when a small set of labelled data is available. A CNN model trained in this way yields an improved accuracy over the state-of-the-art models.
		
	\end{abstract}
	\vspace{-1.0cm}
	\section{Introduction}
	\vspace{-0.4cm}
	
	Incidents of abuse, hate, harassment and misogyny have proliferated with the growing use of social media platforms (e.g. Twitter, Facebook, Instagram). These platforms have generated new opportunities to spread online abuse \cite{dragiewicz2018technology}. The experience of online abuse is a common occurrence for women \cite{amnesty2018toxic}. Often these experiences of online abuse can be categorised as sexist or misogynistic in nature, and can include name-calling and offensive language, threats of harm or sexual violence, intimidation, shaming, and the silencing of women. While it is easy to identify instances of abuse and major abusive campaigns on social media, it is difficult to understand changes in levels of abuse over time, and almost impossible to identify the effectiveness of interventions by platforms in combating abuse\cite{suzor2018evaluating}. An automated system to identify abusive tweets could help in ongoing efforts to develop effective remedies. 
    
	A key challenge in the automatic detection of misogynistic abusive tweets is understanding the context of an individual tweet. We focus here on misogynistic tweets that are abusive towards an individual or group -- a subset of the larger category of tweets that include sexist or misogynistic words or concepts. Accordingly, this study sought to address the difficult task of separating abusive tweets from tweets that were sarcastic, joking, or contained misogynistic keywords in a non-abusive context. A lexical detection approach tends to have low accuracy \cite{davidson2017automated,xiang2012detecting} because they classify all tweets containing particular keywords as misogynistic. Xiang et al. \cite{xiang2012detecting} reported that bag-of-words, part-of-speech (POS) and belief propagation did not work well for the detection of profane tweets because of the significant noise in tweets. For example, tweets do not follow a standard language format, words are often misspelled or altered, and tweets often include words from local dialects or foreign languages. The automated algorithms should look for patterns, sequences, and other complex features that are present, despite the noise, and are correlated with misogynistic tweets. Traditional algorithms (e.g. Random Forest, Logistic Regression and Support Vector Machines) rely on manual process to obtain these kinds of features and are limited by the kinds of features available. Neural Network (NN)-based models, on the other hand, can automatically learn complex features and effectively use them to classify a given instance. 
	
	Relying on a sufficiently large training dataset, CNN models have shown to be effective in Natural Language Processing (NLP) tasks such as semantic parsing \cite{yih2014semantic}, document query matching \cite{shen2014learning}, 
	sentence classification \cite{kim2014convolutional}, etc. 
	A set of convolving filters are applied to local features (e.g. words) 
	to learn patterns similar to $n$Grams. 
	Local features are commonly represented as word vectors where words are projected from a sparse representation onto a lower dimensional vector space. Word vectors essentially encode semantic features of each word in a fixed number of abstract topics (or dimensions).  
	The apparent success of CNN in NLP tasks can be credited to its capability to learn text patterns in semantic space. However, given the requirement of setting the large number of parameters in CNN, often in millions, the CNN models are trained on a huge labelled dataset. In general, this is a limitation of any NN-based model \cite{fadaee2017data}. Overall, curating a large set of labelled tweets containing misogynistic abuse is difficult and costly to achieve due to the large amount of data that needs to be manually examined to rigorously identify abusive tweets. 
	
	However, word vectors trained on a general-purpose corpus cannot capture the task-specific semantics because the nature of general-purpose corpus and the misogynistic tweets is completely different. For example, many words used in these tweets are linguistically specific and unique to Twitter-based discussion, and are not covered in a general-purpose corpus. Consequently, a CNN classifier model built using these word vectors cannot adequately detect misogynistic abusive tweets. 
	
	In this paper, we investigate the effectiveness of various corpus to generate pre-trained word vectors for building a CNN model  when there is a small set of labelled data available. In particular, we trained a CNN using a small set of labelled data to detect misogynistic tweets. We pre-trained word vectors on 0.2 billion unlabelled tweets that contain at least one misogynistic keyword (i.e. \emph{whore, slut, rape}). We customised and regularised the CNN architecture used in \cite{kim2014convolutional}. On the test dataset, the trained CNN model achieves significantly better results than the state-of-the-art models. It is better by a large margin in comparison to the CNN models build on word vectors pre-trained on a sizeable general corpus. The experimental results show that a CNN classifier can be trained on a small labelled tweet data, provided that the word vectors are pre-trained in the context of the problem domain and a careful model customisation and regularisation is performed. 
	
	This project investigates how to effectively apply data mining methods, with a focus on training a NN model, to detect misogynistic tweets. Three main contributions of this paper are: (a) it shows that word vectors pre-trained on a task-specific domain can be used to effectively train CNN when a small set of labelled data is available; (b) it shows how to customise and regularise a CNN architecture to detect misogynistic tweets; and (c) finally, we present an automated data mining method  to detect misogynistic abusive tweets.
%
%
	
	\vspace{-0.5cm}
	\section{Related Work}
	\vspace{-0.4cm}
	Misogynistic abusive tweet detection falls into the research area of text classification. Popular text classification algorithms used in hate speech and offensive language detection are Naive Bayes 
	\cite{davidson2017automated}, Logistic Regression 
	\cite{davidson2017automated},
	Support Vector Machine (SVM) \cite{davidson2017automated,warner2012detecting,xiang2012detecting}
	and Random Forest \cite{davidson2017automated,xiang2012detecting}. Performance of these algorithms depend on feature engineering and feature representation \cite{davidson2017automated,xiang2012detecting}
	. There have been some works where syntactic features are leveraged to identify the targets and the intensity of hate speech. Examples of these features are relevant verb and noun occurrences (e.g. \emph{kill} and \emph{Jews}) \cite{gitari2015lexicon}, 
	and the syntactic structures: I $<$intensity$>$$<$user intent$>$$<$hate target$>$ (e.g. \emph{I f$*$cking hate white people}) \cite{silva2016analyzing}.
	
	Misogynistic tweet detection is challenging for text classification methods because social media users very commonly use offensive words or expletives in their online dialogue \cite{wang2014cursing}. For example, the bag-of-words approach is straightforward and usually has a high recall, but it results in higher number of false positives because the presence of misogynistic words causes these tweets to be misclassified as abusive tweets \cite{kwok2013locate}. 
%
%
	
	Recently, neural network-based classifiers have become popular as they automatically learn abstract features from the given input feature representation \cite{badjatiya2017deep}.
	Input to these algorithms can be various forms of feature encoding, including many of those used in the classic methods. Algorithm design in this category focuses on the selection of the network topology to automatically extract useful abstract features. Popular network architectures are CNN, Recurrent Neural Networks (RNN) and Long Short-Term Memory network (LSTM). CNN is well known for extracting patterns similar to phrases and $n$Grams  \cite{badjatiya2017deep}.
	On the other hand, RNN and LSTM are effective for sequence learning 
	such as order information in text \cite{badjatiya2017deep}. The CNN model has been successfully used for sentence classification \cite{kim2014convolutional} . To effectively identify patterns in the text, they used word embedding pre-trained on Google News corpus while  training a CNN model on the labelled dataset. 

	The neural network-based classifiers have not yet been applied in misogynistic tweet detection. It requires a rigorous investigation as to what extent patterns and orderly information are present in misogynistic tweets, and how we can optimise a Neural Network for classification accuracy. There are many CNN architectures used in the current literature, but the design of an architecture heavily depends on the problem at hand. Therefore, a customised CNN architecture is needed to classify misogynistic tweets. It also remains to be seen whether the word embedding can be sensitive to the domain knowledge of the corpus. Does the word embedding need to be trained on a similar tweet stream to capture contextual properties?

	\vspace{-0.6cm}
	\section{Problem Formulation}
	\label{sec:ProblemFormulation}
	\vspace{-0.5cm}
	Misogynistic abusive tweets may contain misogynistic keywords, but tweets can also be misogynstic abuse without explicitly containing these slurs. Further, not all tweets that contain misogynistic keywords are abusive. Classifying misogynistic abuse in tweets requires close reading, and even humans can struggle to classify these tweets accurately. The focus of this research is to detect abusive tweets that contain misogynistic words. A previous study has identified that three keywords --  \emph{whore}, \emph{slut} and \emph{rape} -- are useful in identifying a substantial portion of misogynistic tweets \cite{bartlett2014misogyny}. However, these misogynistic words are commonly used in tweets that are not abusive, and separating abusive tweets from non-abusive tweets is difficult when we base our classification purely on the occurrence of these words. We propose a two-step method to approach this problem: 
	\vspace{-0.2cm}
	\begin{itemize}
		\item Pre-filtering: We pre-filter tweets that contain any of the three main misogynistic keywords (slut, rape, whore) to find potentially-misogynistic tweets.
		\item Training a CNN model: Using a small labelled data set,  a CNN model is trained to classify the remaining potentially-abusive tweets. We propose several methods to accurately train the model.   
	\end{itemize}
	\vspace{-0.2cm}
	The research team used a systematic approach to generate the labelled data manually. The following contextual information was used in assessing whether a tweet contains targeted misogynistic abuse, or not: 
	(a) Is a specific person or group being targeted in this tweet?
	(b) Does this tweet contain a specific threat or wish for violence?
	(c) Does this tweet encourage or promote self-harm or suicide?
	(d) Is the tweet harassing a specific person, or inciting others to harass a specific person?
	(e) Does the tweet use misogynistic language in objectifying a person, making sexual advances, or sending sexually explicit material?
	(f) Is the tweet promoting hateful conduct by gender, sexual orientation, etc.?
	
	The labelled tweets reveal many challenges that need to be addressed to train a classifier effectively. These included:
	(a) The misogynistic words are not the discriminatory words. Many keywords are overlapping between misogynistic and non-misogynistic tweets, especially misogynistic keywords.
	(b) Words may be misspelt and spelt in many ways.
	(c) Sometimes people mix words from local dialects or foreign languages.
	(d) The  data is noisy and does not follow a standard language sequence (format).
	(e) Effectively detecting misogynistic tweets needs access to semantics and context information that is often not available, e.g. it is difficult to use dictionary-based semantics for the nature of noise in tweets and difficult to know the context because of the small length of a tweet. 
	(f) The labelling process is time consuming and it is extremely difficult to generate a large quantity of labelled data because only a very small portion of tweets can be identified as misogynistic. 
	
	Given these challenges, in this paper, we investigate how to effectively train a CNN model with a small set of labelled data to detect misogynistic abusive tweets. We train a CNN on top of the pre-trained word vectors (a.k.a. word embedding or distributed representation of words). The primary focus is to find out what kind of pre-trained word vectors is useful to train a CNN with a small dataset. Another two important focuses are to find out what customised architecture of CNN is effective in the given problem and to test the effectiveness of some simple data and feature augmentation. 
	
	\vspace{-0.7cm}
	\section{Word Embedding}
\begin{figure}[htb]
		\centering
		\scriptsize
		\includegraphics[width=0.6\textwidth]{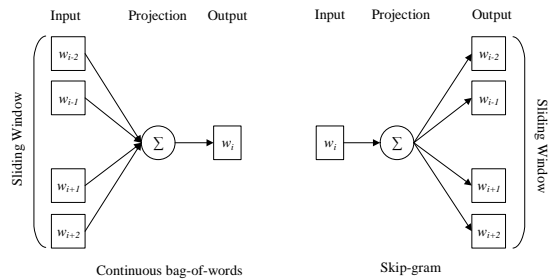}
		\caption{Word Embedding Models}
		\label{fig:Word2vec_Architectures}
	\end{figure}

	\vspace{-0.5cm}
	Word embedding models map each word from the vocabulary to a vector of real numbers. They aim to quantify and categorise semantic similarities between words based on their distributional property based on the premise that a word is characterised by the company it keeps. Given a sizeable unlabelled corpus, these models can effectively learn a high-quality word embedding. Based on the feed-forward neural network, Mikolov et al. \cite{mikolov2013distributed} proposed two popular models: Skip-gram and Continuous Bag-of-Words as shown in Figure \ref{fig:Word2vec_Architectures}.
	
	Given the words within a sliding window, the continuous bag-of-words model predicts the current word $w_i$ from the surrounding context words $C$, i.e. $p(w_i|C)$.  In contrast, the skip-gram model uses the current word $w_i$ to predict the surrounding context words $C$, i.e. $p(C|w_i)$. In Figure \ref{fig:Word2vec_Architectures}, for example, if the current position of a running sliding window contains the phrase \emph{she looks like a crack whore}. In continuous bag-of-words, the context words \{she, looks, like, a, whore\} can be used to predict the current word \{crack\}, whereas, in skip-gram, the current word \{crack\} can be used to predict the context words \{she, looks, like, a, whore\}. 
	
	\vspace{-0.5cm}
	\vspace{-0.5cm}
	
	The training objective is to find a word embedding that maximises $p(t_i|C)$ or $p(C|t_i)$ over a training dataset. In each step of training, each word is either (a) pulled closer to the words that co-occur with it or (b) pushed away from all the words that do not co-occur with it. A\textit{ softmax} or \textit{approximate softmax} function can be used to achieve this objective \cite{mikolov2013distributed}. At the end of the training, the embedding brings closer not only the words that are explicitly co-occurring in a training dataset, but also the words that implicitly co-occur. For example, if $t_1$ explicitly co-occurs with $t_2$ and $t_2$ explicitly co-occurs with $t_3$, then the model can bring closer not only $t_1$ to $t_2$, but also $t_1$ to $t_3$. The continuous bag-of-words model is faster 
	and has slightly better accuracy for the words that appear frequently. 
	Therefore, we use this model in this research.  
	
	\vspace{-0.4cm}
	\section{Model Architecture}
\begin{figure}[htb]
		\centering
		\scriptsize
		\includegraphics[width=0.7\textwidth]{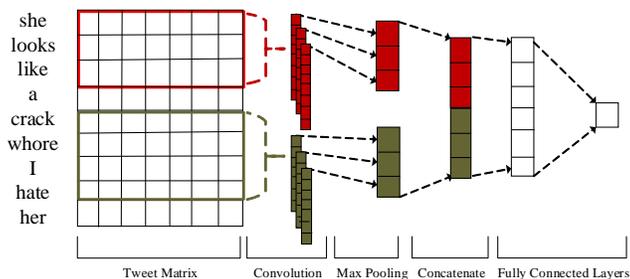}
		\caption{CNN Model Architecture}
		\label{fig:ModelArchitecture}
	\end{figure}

	\vspace{-0.4cm}
	We empirically customise and regulate Kim's \cite{kim2014convolutional} CNN architecture to detect misogynistic tweets and reduce overfitting. Figure \ref{fig:ModelArchitecture} shows the architecture.  
	We use word embedding to represent each word $w$ in an $n$-dimensional word vector $\mathbf{w} \in \mathbb{R}^n$. A tweet $t$ with $m$ words is represented as a matrix $\mathbf{t} \in \mathbb{R}^{m \times n}$. Convolution operation is applied to the tweet matrix with one stride. Each convolution operation applies a filter $\mathbf{f}_i \in \mathbb{R}^{h \times n}$ of size $h$. Empirically, based on the accuracy improvement in ten-fold cross validation, we used 256 filters for $h \in \{3,4\}$ and 512 filters for $h \in \{5\}$. The convolution is a function $\mathbf{c}(\mathbf{f}_i, \mathbf{t}) = r(\mathbf{f}_i \cdot \mathbf{t}_{k:k+h-1})$, where $\mathbf{t}_{k:k+h-1}$ is the $k$th vertical slice of the tweet matrix from position $k$ to $k+h-1$, $\mathbf{f}_i$ is the given filter and $r$ is a ReLU function. The function $\mathbf{c}(\mathbf{f}_i, \mathbf{t})$ produces a feature $c_k$ similar to $n$Grams or phrases for each slice $k$, resulting in $m-n+1$ features. We apply the max-pooling operation over these features and take the maximum value, i.e. $ \hat{c}_i = \max\mathbf{c}(\mathbf{f}_i, \mathbf{t})$. Max-pooling is carried to capture the most important feature for each filter. As there are a total of 1024 filters (256+256+512) in the proposed model, the 1024 most important features are learned from the convolution layer. 
	
	These features are passed to a fully connected hidden layer with 256 perceptrons that use the ReLU activation function. This fully connected hidden layer allows learning the complex non-linear interactions between the features from the convolution layer and generates 256 higher level new features. Finally these 256 higher level features are passed to the output layer with single perceptron that uses the sigmoid activation function. The perceptron in this layer generates the probability of the tweet being misogynistic. 
	
	We randomly dropout a proportion of units from each layer except the output layer by setting them to zero. This is done to prevent co-adaptation of units in a layer and to reduce overfitting. We empirically dropout 50\% units from the input layer, the filters of size 3 and the fully connected hidden layer. We dropout only 20\% units from the filters of size 4 and 5.  
	
	\vspace{-0.7cm}
	\setlength{\textfloatsep}{0pt}

	\vspace{-1.4cm}
	\section{Empirical evaluation}
	\vspace{-0.5cm}
	The primary objectives of the experiments are to show: (a) word vectors pre-trained on a task-specific domain is more effective than those pre-trained on a sizeable general corpus; (b) CNN trained on a small dataset and built on word vectors pre-trained on a task-specific domain can perform better than the state-of-the-art models; and (c) the impact of some simple data and word augmentation techniques on training a CNN model.
	
	\vspace{-0.6cm}
	\subsection{Data Collection}

	\vspace{-0.2cm}
	\subsubsection{Labelled Tweets:}
	We collected tweets using Twitter's streaming API. For the labelled dataset, we identified 10k tweets that contain any of the three main misogynistic keywords (i.e., whore, slut, rape). Following the misogynistic tweet definition in Section \ref{sec:ProblemFormulation}, the research team labelled a total of 5000 tweets with 1800 misogynistic and 3200 non-misogynistic labels. A stratified data selection was made to reduce a trained models' bias to a specific label, i.e. we kept 1800 misogynistic and 1800 randomly selected nonmisogynistic tweets. We used 80\% examples for training and 20\% for testing. We used the ten-fold cross-validation to tune hyperparameters. We used the Porter's suffix-stripping algorithm for preprocessing.
	
	The tweet labelling method has the following limitations: 
	(a) The coding is based on a literal interpretation of the text; with limited context, we are likely to include some sarcasm or humour.
	(b) We are only labelling tweets written in English.
	(c) Identifying the tweets by keywords only, we will not catch abuse that appears to be ordinary misogyny, e.g. \emph{get back in the kitchen}.
	(d) Identifying the tweets by keywords only, we will not identify harassment that is targeted and organised harassment, either ongoing over time or involving many participants, but does not use one of our keywords.
	
	\vspace{-0.6cm}
	\subsubsection{WikiNews:}
	Word vectors of 300-dimension pre-trained on the Wikipedia 2017, UMBC webbase corpus and statmt.org news datasets containing a total of 16 billion words using fastText (a library for learning word embeddings created by Facebook's AI Research lab) \cite{mikolov2018advances}.
	
	\vspace{-0.7cm}
	\subsubsection{GoogleNews:}
	Word vectors of 300-dimension pre-trained on Google News corpus\footnote{https://bit.ly/2esteWf} containing a total of three billion words using the Continuous Bag-of-Words Word2vec model \cite{mikolov2013efficient}.
	
	\vspace{-0.7cm}
	\subsubsection{Potentially Misogynistic Tweets:}
	Word vectors of 200-dimension pre-trained on 0.2 billion tweets that contain any of the three main misogynistic keywords. A Continuous Bag-of-Words Word2vec model is used in pre-training while minimum count for word is set to 100.
	
	\vspace{-0.6cm}
	\subsection{Evaluation Measures}
	\vspace{-0.4cm}
	We used six standard evaluation measures of classification performance:  Accuracy, Precision, Recall, F$_1$ Score, Cohen Kappa (CK) and Area Under Curve (AUC). We also report True Positive (TP), True Negative (TN), False Positive (FP) and False Negative (FN) values.
	
	\vspace{-0.4cm}
	\subsection{Baseline Models}
	\vspace{-0.4cm}
	We have implemented eight baseline models to compare the performance with the proposed CNN model.
	\vspace{-0.2cm}
	\begin{itemize}
		\item Long Short-Term Memory Network (LSTM) \cite{hochreiter1997long}. We have implemented LSTM with 100 units, 50\% dropout, binary cross-entropy loss function, Adam optimiser and sigmoid activation. 
		\item Feedforward Deep Neural Network (DNN) \cite{glorot2010understanding}. We have implemented DNN with five hidden layers, each layer containing eight units, 50\% dropout applied to the input layer and the first two hidden layers, softmax activation and 0.04 learning rate. For all neural network based models (CNN, LSTM, DNN), hyperparameters are manually tuned based on ten-fold cross-validation. 
        \item Non NN models including Support Vector Machines (SVM) \cite{hearst1998support},  Random Forest \cite{liaw2002classification}, XGBoost (XGB) \cite{chen2016xgboost},
		Multinomial Naive Bayes (MNB) \cite{lewis1998naive}, k-Nearest Neighbours (kNN) \cite{weinberger2009distance} and Ridge Classifier (RC) \cite{hoerl1970ridge}. Hyperparameters of all these models are automatically tuned using ten-fold cross-validation and GridSearch from sklearn. 
	\end{itemize}
	\vspace{-0.8cm}
	
	\subsection{Results and Discussion}
	\vspace{-0.2cm}
	
	\subsubsection{Word Embedding Performances}
	We conducted experiments to see the effects of different word embeddings in training the CNN model. A summary of the embeddings is given in Table \ref{tab:WESummary} and the experimental results are given in Table \ref{tab:WEResults}. Three main observations from the results are: (a) Word vectors pre-trained on a large dataset (e.g., WE1, WE2, WE4, WE5) always improves performance. The convolution layer, that captures $n$Gram-like patterns in the tweets while using word vectors to represent the tweets, allows the model to find these patterns in semantic space. The pre-trained word embedding can provide the semantics of words that have fewer appearances in the training dataset. This reinforces the prior finding  \cite{sharif2014cnn} that the features obtained from a pre-trained deep learning model perform well on a variety of tasks. (b) Updating the word vectors with the labelled data while training the classifier improves the performance (e.g., WE1 over WE2). This allows the semantics of words to be more focused over the training set. (c) Word vectors pre-trained on potentially misogynistic tweets and updated with labelled data performs the best. It improves the CNN accuracy by around 12\% compared with word vectors pre-trained on a standard corpus (e.g. Google News corpus). This observation challenges the previous findings \cite{kim2014convolutional} that general pre-trained word vectors (e.g. word vectors pre-trained on Google News) are \emph{universal} feature extractors. Due to the small labelled dataset used in training the CNN model, it was not enough to update the necessary word vectors for the problem domain, given that tweets are very noisy and mostly different from standard corpora like Google News or Wikipedia. The word vectors pre-trained on unlabelled datasets in the task-specific domain can address this problem.
	
	The apparent performance correlation of CNN and word vector can be related to the similar goal that CNN and word vector have.
	In the word vector representation, 
	semantically similar words are represented with similar vectors and semantically dissimilar words are represented with dissimilar vectors. This is obtained through training the word vectors on a corpus where it searches for co-occurring words through a filter called sliding window. 
	To train CNN with a labelled tweet, the words in the tweet are represented with word vectors. CNN discovers patterns in these word vectors through varying length filters, where a pattern identifies something similar to a set of words that co-occur in the labelled tweets. It can be ascertained that both CNN and word vector capture the patterns of co-occurring words. 
	Because CNN learns the patterns in the vector space, it harnesses the patterns (or semantic relations) already learned in the vector space. 
	Thus, pre-trained word vectors, especially trained on a corpus from the similar nature domain, may significantly improve the performance of CNN model when only a small labeled dataset is available for training. Pre-trained word vectors acts as the smoothing used in many language models \cite{zhai2017study}.

	\vspace{-0.8cm}
	\setlength{\textfloatsep}{0pt}
	\begin{figure}[htb!]
		\centering
		\begin{minipage}{0.54\linewidth}%
			\centering
			\scriptsize
			\caption{Summary of Word Embeddings}
			\begin{tabular}{p{.7cm}|p{5.5cm}}
				\toprule
				Model & Description \\
				\midrule
				WE1   & W2V pre-trained on potentially abusive tweets and updated with labelled data \\
				\midrule
				WE2   & W2V pre-trained on potentially abusive tweets but not updated with labelled data \\
				\midrule
				WE3   & W2V Trained with only labelled data \\
				\midrule
				WE4   & W2V pre-trained on google news and updated with labelled data \\
				\midrule
				WE5   & fastText pre-trained on Wikipedia pages and updated with labelled data \\
				\bottomrule
			\end{tabular}%
			\label{tab:WESummary}%
		\end{minipage}
		\begin{minipage}{0.42\linewidth}%
			\centering
			\scriptsize
			\caption{Performance of CNN applied on different Word Embeddings}
			\begin{tabular}{lccccc}
				\toprule
				& WE1 \quad & WE2 \quad & WE3 \quad & WE4 \quad & WE5 \\
				\midrule
				TP & \textbf{267}   & 264   & 194   & 217   & 199 \\
				TN & \textbf{283}   & 279   & 273   & 274   & 281 \\
				FP & \textbf{78}    & 82    & 88    & 87    & 80 \\
				FN & \textbf{94}    & 97    & 167   & 144   & 162 \\
				Accuracy & \textbf{0.762} & 0.752 & 0.647 & 0.680 & 0.665 \\
				Precision &\textbf{ 0.774} & 0.763 & 0.688 & 0.714 & 0.713 \\
				Recall & \textbf{0.740} & 0.731 & 0.537 & 0.601 & 0.551 \\
				F$_1$ Score &\textbf{ 0.756} & 0.747 & 0.603 & 0.653 & 0.622 \\
				CK & \textbf{0.524} & 0.504 & 0.294 & 0.360 & 0.330 \\
				AUC & \textbf{0.762} & 0.752 & 0.647 & 0.680 & 0.665 \\
				\bottomrule
			\end{tabular}%
			\label{tab:WEResults}%
		\end{minipage}
	\end{figure}
	
    \vspace{-1.3 cm}
	\subsubsection{Classifier Models Comparison}
	We implemented the proposed CNN model and the eight baseline models to detect misogynistic tweets. Guided by the experimental results in previous section, both CNN and LSTM models were built on word vectors that are pre-trained on potentially abusive tweets and updated with the labelled dataset during the classifier training. Performances of the models are summarised in Table \ref{tab:ClassifierPerformance}.  
	
	Result shows that CNN outperforms all other models. For example, the improvement in precision, accuracy,  Cohen Kappa score and AUR of CNN over the second best performing model LSTM are 6.120\%, 4.364\%, 13.855\% and 4.364\% respectively. LSTM is known to be effective in text datasets and the results reflect this. The reason for CNN outperforming LSTM and other baseline models might be the nature of tweets. Tweets are super condensed texts, full of noise and often do not follow the standard sequence of the language. Traditional models (e.g. RF, SVC, kNN, etc.) are based on bag-of-words representation that can be highly impacted by the significant noise in tweets  \cite{xiang2012detecting}. Besides, the bag-of-words representation cannot capture sequences and patterns that are very important to identify a misogynistic tweet. For example, if a tweet contains a sequence \emph{if you know what I mean}, there is a high chance that this tweet might be misogynistic, even though individual keywords are innocent. 
%
%
	
	The performance of LSTM is better than traditional models as it can capture sequences. However, sequences in tweets often get altered by noises (e.g. misspelled or intentionally altered by the author); therefore LSTM might struggles to detect misogynistic tweets.  
	CNN models are well known for effectively discovering a large number of patterns and sub-patterns through many filters with varying size. If a few words of a given tweet are altered by noise it can still match a sub-pattern. This means CNN is less affected by noise. 
	As a result CNN out performs LSTM. 
	
	CNN is popularly used in Computer Vision and is known to be effective only if the model is trained on massive datasets. However, in this research, we trained a simple CNN with only three thousand labelled tweets. This simple CNN uses only one layer of convolutions on top of word vectors, and it achieves significantly better results than state-of-the-art models. These results ascertain that a CNN can be trained on a small labelled dataset, provided that word vectors are pre-trained in the context of the problem domain, and a careful model customisation and some regularisations are performed. 
	
	\vspace{-0.7cm}
	\setlength{\textfloatsep}{0pt}
	\begin{table}[htbp]
		\centering
		\scriptsize
		\caption{Performances of Classification Models}
		\vspace{-4mm}
		\begin{tabular}{lccccccccc}
			\toprule
			& CNN \quad \quad & LSTM \quad \quad & DNN \quad  \quad & SVC \quad \quad  & RF \quad \quad & XGB \quad \quad & MNB \quad \quad & kNN \quad \quad & RC \\
			\midrule
			TP & \textbf{267}   & 264   & 275   & 257   & 279   & 286   & 272   & 95    & 263 \\
			TN & \textbf{283}   & 263   & 171   & 244   & 229   & 223   & 251   & 302   & 245 \\
			FP & \textbf{78}    & 98    & 190   & 117   & 132   & 138   & 110   & 59    & 116 \\
			FN & \textbf{94}    & 97    & 86    & 104   & 82    & 75    & 89    & 266   & 98 \\
			Accuracy & \textbf{0.762} & 0.730 & 0.618 & 0.694 & 0.704 & 0.705 & 0.724 & 0.550 & 0.704 \\
			Precision & \textbf{0.774} & 0.729 & 0.591 & 0.687 & 0.679 & 0.675 & 0.712 & 0.617 & 0.694 \\
			Recall & \textbf{0.740} & 0.731 & 0.762 & 0.712 & 0.773 & 0.792 & 0.753 & 0.263 & 0.729 \\
			F$_1$ Score & \textbf{0.756} & 0.730 & 0.666 & 0.699 & 0.723 & 0.729 & 0.732 & 0.369 & 0.711 \\
			CK & \textbf{0.524} & 0.460 & 0.235 & 0.388 & 0.407 & 0.410 & 0.449 & 0.100 & 0.407 \\
			AUC & \textbf{0.762} & 0.730 & 0.618 & 0.694 & 0.704 & 0.705 & 0.724 & 0.550 & 0.704 \\
			\bottomrule
		\end{tabular}%
		\label{tab:ClassifierPerformance}%
	\end{table}%
	\vspace{-1.3cm}
	
	\subsubsection{Data and Word Augmentation Performances}
	Data augmentation and document expansion is popularly used in computer vision and information retrieval respectively to artificially inflating a small labelled dataset and/or input vectors. In this paper, we augmented/expanded the data multiple ways and studied their  impact on training the CNN model. We used two sources of data to generate augmented data: (1) the word vectors pre-trained on the potential misogynistic tweets; and (2) topics identified by Non-Negative Matrix Factorisation (NMF) \cite{lee2001algorithms} on the tweet training dataset, performed separately on each class. A total of six policies were  followed.
	AT1: Words in a labelled tweet are randomly replaced by semantically similar words from word vector space to create an artificial tweet.
	AT2: Discriminative Words in a labelled tweet are randomly replaced by semantically similar words from word vector space to create an artificial tweet. A discriminative word is a word that more frequently appears in the tweet of a specific label.  
	AT3: A tweet is expanded by adding its semantically similar words found from word vector space.
	AT4: A tweet is expanded by adding its semantically similar words found from NMF.
	AT5: Use the discovered topics in NMF as artificial tweets.
	AT6: A set of words from word vector space that is semantically similar to a tweet is used as an artificial tweet.
	
	Table \ref{tab:DapResults} reports the performance of model trained with each of these augmentation policies and the CNN model trained with the original labelled dataset before any augmentation (labelled as AT0). The experimental results show that these ways of augmentation do not improve the accuracy. We conjecture that additional external features (i.e. words) may distort the patterns exist in the original tweets, since the CNN classifier largely depends on learning these patterns, the performance degrades.

	\vspace{-0.5cm}
	\setlength{\textfloatsep}{0pt}
	\begin{table}[htbp]
		\centering
		\scriptsize
		\caption{CNN results from Data Augmentation Policies}
		\vspace{-4mm}
		\begin{tabular}{lccccccc}
			\toprule
			& AT0 \quad \quad & AT1 \quad \quad & AT2 \quad \quad & AT3 \quad \quad & AT4 \quad \quad & AT5 \quad \quad & AT6 \\
			\midrule
			TP & 267   & 270   & 242   & 301   & 281   & 292   & 288 \\
			TN & 283   & 275   & 287   & 224   & 203   & 238   & 241 \\
			FP & 78    & 86    & 74    & 137   & 158   & 123   & 120 \\
			FN & 94    & 91    & 119   & 60    & 80    & 69    & 73 \\
			Accuracy & 0.762 & 0.755 & 0.733 & 0.727 & 0.670 & 0.734 & 0.733 \\
			Precision & 0.774 & 0.758 & 0.766 & 0.687 & 0.640 & 0.704 & 0.706 \\
			Recall & 0.740 & 0.748 & 0.670 & 0.834 & 0.778 & 0.809 & 0.798 \\
			F$_1$ Score & 0.756 & 0.753 & 0.715 & 0.753 & 0.703 & 0.753 & 0.749 \\
			CK & 0.524 & 0.510 & 0.465 & 0.454 & 0.341 & 0.468 & 0.465 \\
			AUC & 0.762 & 0.755 & 0.733 & 0.727 & 0.670 & 0.734 & 0.733 \\
			\bottomrule
		\end{tabular}%
		\label{tab:DapResults}%
	\end{table}%

	\section{Conclusions}
	\vspace{-0.4cm}
	This paper presents a novel method of misogynistic tweet detection using word embedding and the CNN model when only a small amount of labelled data is available. We report the results of a series of experiments conducted to investigate the effectiveness of training a model with a small dataset. We customised and regularised a CNN architecture, and it performs better than the state-of-the-art models, provided that the CNN is built on word vectors pre-trained on the task-specific domain. Experimental results show that a CNN model built on word vectors pre-trained on the task-specific unlabelled dataset is more effective than built on word vectors pre-trained on a sizeable general corpus. Experimental results also show that simple data augmentation policies are not adequate to improve misogynistic tweet detection performance in the CNN model.
	
	\vspace{-0.6cm}
	\section{Acknowledgement}
	\vspace{-0.4cm}
	This research was fully supported by the QUT IFE Catapult fund. Suzor is the recipient of an Australian Research Council DECRA Fellowship (project number DE160101542).
	
	\vspace{-0.5cm}
	\bibliographystyle{spmpsci}
	\bibliography{References}

\begin{thebibliography}{10}
\providecommand{\url}[1]{{#1}}
\providecommand{\urlprefix}{URL }
\expandafter\ifx\csname urlstyle\endcsname\relax
  \providecommand{\doi}[1]{DOI~\discretionary{}{}{}#1}\else
  \providecommand{\doi}{DOI~\discretionary{}{}{}\begingroup
  \urlstyle{rm}\Url}\fi

\bibitem{badjatiya2017deep}
Badjatiya, P., Gupta, S., Gupta, M., Varma, V.: Deep learning for hate speech
  detection in tweets.
\newblock In: Proceedings of the 26th International Conference on World Wide
  Web Companion, pp. 759--760. International World Wide Web Conferences
  Steering Committee (2017)

\bibitem{bartlett2014misogyny}
Bartlett, J., Norrie, R., Patel, S., Rumpel, R., Wibberley, S.: Misogyny on
  twitter.
\newblock Demos  (2014)

\bibitem{chen2016xgboost}
Chen, T., Guestrin, C.: Xgboost: A scalable tree boosting system.
\newblock In: Proceedings of the 22nd acm sigkdd international conference on
  knowledge discovery and data mining, pp. 785--794. ACM (2016)

\bibitem{davidson2017automated}
Davidson, T., Warmsley, D., Macy, M., Weber, I.: Automated hate speech
  detection and the problem of offensive language.
\newblock arXiv preprint arXiv:1703.04009  (2017)

\bibitem{dragiewicz2018technology}
Dragiewicz, M., Burgess, J., Matamoros-Fern{\'a}ndez, A., Salter, M., Suzor,
  N.P., Woodlock, D., Harris, B.: Technology facilitated coercive control:
  domestic violence and the competing roles of digital media platforms.
\newblock Feminist Media Studies pp. 1--17 (2018)

\bibitem{fadaee2017data}
Fadaee, M., Bisazza, A., Monz, C.: Data augmentation for low-resource neural
  machine translation.
\newblock In: Proceedings of the 55th Annual Meeting of the Association for
  Computational Linguistics (Volume 2: Short Papers), vol.~2, pp. 567--573
  (2017)

\bibitem{gitari2015lexicon}
Gitari, N.D., Zuping, Z., Damien, H., Long, J.: A lexicon-based approach for
  hate speech detection.
\newblock International Journal of Multimedia and Ubiquitous Engineering
  \textbf{10}(4), 215--230 (2015)

\bibitem{glorot2010understanding}
Glorot, X., Bengio, Y.: Understanding the difficulty of training deep
  feedforward neural networks.
\newblock In: Proceedings of the thirteenth international conference on
  artificial intelligence and statistics, pp. 249--256 (2010)

\bibitem{hearst1998support}
Hearst, M.A., Dumais, S.T., Osuna, E., Platt, J., Scholkopf, B.: Support vector
  machines.
\newblock IEEE Intelligent Systems and their applications \textbf{13}(4),
  18--28 (1998)

\bibitem{hochreiter1997long}
Hochreiter, S., Schmidhuber, J.: Long short-term memory.
\newblock Neural computation \textbf{9}(8), 1735--1780 (1997)

\bibitem{hoerl1970ridge}
Hoerl, A.E., Kennard, R.W.: Ridge regression: applications to nonorthogonal
  problems.
\newblock Technometrics \textbf{12}(1), 69--82 (1970)

\bibitem{amnesty2018toxic}
International, A.: Toxic twitter - a toxic place for women  (2018).
\newblock \urlprefix\url{https://bit.ly/2FZYQhV}

\bibitem{kim2014convolutional}
Kim, Y.: Convolutional neural networks for sentence classification.
\newblock arXiv preprint arXiv:1408.5882  (2014)

\bibitem{kwok2013locate}
Kwok, I., Wang, Y.: Locate the hate: Detecting tweets against blacks.
\newblock In: AAAI (2013)

\bibitem{lee2001algorithms}
Lee, D.D., Seung, H.S.: Algorithms for non-negative matrix factorization.
\newblock In: Advances in neural information processing systems, pp. 556--562
  (2001)

\bibitem{lewis1998naive}
Lewis, D.D.: Naive (bayes) at forty: The independence assumption in information
  retrieval.
\newblock In: European conference on machine learning, pp. 4--15. Springer
  (1998)

\bibitem{liaw2002classification}
Liaw, A., Wiener, M., et~al.: Classification and regression by randomforest.
\newblock R news \textbf{2}(3), 18--22 (2002)

\bibitem{mikolov2013efficient}
Mikolov, T., Chen, K., Corrado, G., Dean, J.: Efficient estimation of word
  representations in vector space.
\newblock International Conference on Learning Representations (ICLR) Workshop
  (2013)

\bibitem{mikolov2018advances}
Mikolov, T., Grave, E., Bojanowski, P., Puhrsch, C., Joulin, A.: Advances in
  pre-training distributed word representations.
\newblock In: Proceedings of the International Conference on Language Resources
  and Evaluation (LREC 2018) (2018)

\bibitem{mikolov2013distributed}
Mikolov, T., Sutskever, I., Chen, K., Corrado, G.S., Dean, J.: Distributed
  representations of words and phrases and their compositionality.
\newblock In: Advances in neural information processing systems, pp. 3111--3119
  (2013)

\bibitem{sharif2014cnn}
Sharif~Razavian, A., Azizpour, H., Sullivan, J., Carlsson, S.: Cnn features
  off-the-shelf: an astounding baseline for recognition.
\newblock In: Proceedings of the IEEE conference on computer vision and pattern
  recognition workshops, pp. 806--813 (2014)

\bibitem{shen2014learning}
Shen, Y., He, X., Gao, J., Deng, L., Mesnil, G.: Learning semantic
  representations using convolutional neural networks for web search.
\newblock In: Proceedings of the 23rd International Conference on World Wide
  Web, pp. 373--374. ACM (2014)

\bibitem{silva2016analyzing}
Silva, L.A., Mondal, M., Correa, D., Benevenuto, F., Weber, I.: Analyzing the
  targets of hate in online social media.
\newblock In: ICWSM, pp. 687--690 (2016)

\bibitem{suzor2018evaluating}
Suzor, N., Van~Geelen, T., Myers~West, S.: Evaluating the legitimacy of
  platform governance: A review of research and a shared research agenda.
\newblock International Communication Gazette \textbf{80}(4), 385--400 (2018)

\bibitem{wang2014cursing}
Wang, W., Chen, L., Thirunarayan, K., Sheth, A.P.: Cursing in english on
  twitter.
\newblock In: Proceedings of the 17th ACM conference on Computer supported
  cooperative work \& social computing, pp. 415--425. ACM (2014)

\bibitem{warner2012detecting}
Warner, W., Hirschberg, J.: Detecting hate speech on the world wide web.
\newblock In: Proceedings of the Second Workshop on Language in Social Media,
  pp. 19--26. Association for Computational Linguistics (2012)

\bibitem{weinberger2009distance}
Weinberger, K.Q., Saul, L.K.: Distance metric learning for large margin nearest
  neighbor classification.
\newblock Journal of Machine Learning Research \textbf{10}(Feb), 207--244
  (2009)

\bibitem{xiang2012detecting}
Xiang, G., Fan, B., Wang, L., Hong, J., Rose, C.: Detecting offensive tweets
  via topical feature discovery over a large scale twitter corpus.
\newblock In: Proceedings of the 21st ACM international conference on
  Information and knowledge management, pp. 1980--1984. ACM (2012)

\bibitem{yih2014semantic}
Yih, W.t., He, X., Meek, C.: Semantic parsing for single-relation question
  answering.
\newblock In: Proceedings of the 52nd Annual Meeting of the Association for
  Computational Linguistics (Volume 2: Short Papers), vol.~2, pp. 643--648
  (2014)

\bibitem{zhai2017study}
Zhai, C., Lafferty, J.: A study of smoothing methods for language models
  applied to ad hoc information retrieval.
\newblock In: ACM SIGIR Forum, vol.~51, pp. 268--276. ACM (2017)

\end{thebibliography}

\end{document}